%% file: main.tex
\documentclass[10pt,twocolumn,letterpaper]{article}
\usepackage{cvpr}
\input{preamble}

\definecolor{cvprblue}{rgb}{0.21,0.49,0.74}
\usepackage[pagebackref,breaklinks,colorlinks,citecolor=cvprblue]{hyperref}
\usepackage{algorithm}
\usepackage{algpseudocode}
\usepackage{amsfonts}
\usepackage{graphicx}
\usepackage{float}
\usepackage{colortbl}
\usepackage{mathtools}
\usepackage{multicol}
\usepackage{multirow}

\title{Dexterous Grasp Transformer}

\begin{document}

{\onecolumn
\noindent \vspace{1cm}

\noindent \textbf{\huge\centering{Dexterous Grasp Transformer}}

\vspace{2cm}

\noindent {\LARGE{Guo-Hao Xu\textsuperscript{*}, Yi-Lin Wei\textsuperscript{*}, \\
Dian Zheng, Xiao-Ming Wu, Wei-Shi Zheng\textsuperscript{\dag}}}
\\
\\
\textsuperscript{*}Equal contribution \\
\textsuperscript{\dag}Corresponding author: Wei-Shi Zheng.
\\
\\
Code: \href{https://github.com/iSEE-Laboratory/DGTR}{\textcolor{cvprblue}{https://github.com/iSEE-Laboratory/DGTR}}
\\
\\
Project page: \href{https://isee-laboratory.github.io/dgtr/}{\textcolor{cvprblue}{https://isee-laboratory.github.io/dgtr/}}

\vspace{1cm}

\noindent {\LARGE{Accepted date: 27-Feb-2024 to  IEEE/CVF Conference on Computer Vision and Pattern Recognition}}

\vspace{1cm}

\noindent For reference of this work, please cite:

\vspace{1cm}
\noindent Guo-Hao Xu, Yi-Lin Wei, Dian Zheng, Xiao-Ming Wu, and Wei-Shi Zheng.~Dexterous Grasp Transformer.~~In \emph{Proceedings of the IEEE Conference on
Computer Vision and Pattern Recognition,} 2024.

\vspace{1cm}

\noindent Bib:\\
\noindent @inproceedings\{xu2024dexterous,\\
\indent title = \{Dexterous Grasp Transformer\}, \\
\indent author = \{Xu, Guo-Hao and Wei, Yi-Lin and Zheng, Dian and Wu, Xiao-Ming and Zheng, Wei-Shi\},\\
\indent booktitle = \{Proceedings of the IEEE/CVF Conference on Computer Vision and Pattern Recognition\},\\
\indent year = \{2024\}\\
\}
}
\twocolumn

\author{
    Guo-Hao Xu\footnotemark[1] \textsuperscript{ 1},
    \quad Yi-Lin Wei\footnotemark[1] \textsuperscript{ 1},
    \quad Dian Zheng \textsuperscript{1},
    \quad Xiao-Ming Wu \textsuperscript{1},
    \quad Wei-Shi Zheng \footnotemark[2] \textsuperscript{ 1,2} \\
    \textsuperscript{1} School of Computer Science and Engineering, Sun Yat-sen University, China \\
    \textsuperscript{2} Key Laboratory of Machine Intelligence and Advanced Computing, Ministry of Education, China \\
    \tt\small {
        \{xugh23, weiylin5, zhengd35, wuxm65\}@mail2.sysu.edu.cn
        \quad wszheng@ieee.org
    }
}
\maketitle
\renewcommand{\thefootnote}{\fnsymbol{footnote}}
\footnotetext[1]{Equal contribution.}
\footnotetext[2]{Corresponding author.}
\begin{abstract}
In this work, we propose a novel discriminative framework for dexterous grasp generation, named \textbf{D}exterous \textbf{G}rasp \textbf{TR}ansformer (\textbf{DGTR}), capable of predicting a diverse set of feasible grasp poses by processing the object point cloud with only \textbf{one forward pass}. We formulate dexterous grasp generation as a set prediction task and design a transformer-based grasping model for it. However, we identify that this set prediction paradigm encounters several optimization challenges in the field of dexterous grasping and results in restricted performance. To address these issues, we propose progressive strategies for both the training and testing phases. First, the dynamic-static matching training (DSMT) strategy is presented to enhance the optimization stability during the training phase. Second, we introduce the adversarial-balanced test-time adaptation (AB-TTA) with a pair of adversarial losses to improve grasping quality during the testing phase. Experimental results on the DexGraspNet dataset demonstrate the capability of DGTR to predict dexterous grasp poses with both high quality and diversity. Notably, while keeping high quality, the diversity of grasp poses predicted by DGTR significantly outperforms previous works in multiple metrics without any data pre-processing.
Codes are available at \href{https://github.com/iSEE-Laboratory/DGTR}{https://github.com/iSEE-Laboratory/DGTR}.
\end{abstract}

\section{Introduction}
Robotic dexterous grasping stands as a fundamental and critical task in the field of robotics and computer vision, offering a versatile and fine-grained approach with extensive applications in industrial production and daily scenarios.

\begin{figure}[t]
\centering
\includegraphics[width=\linewidth]{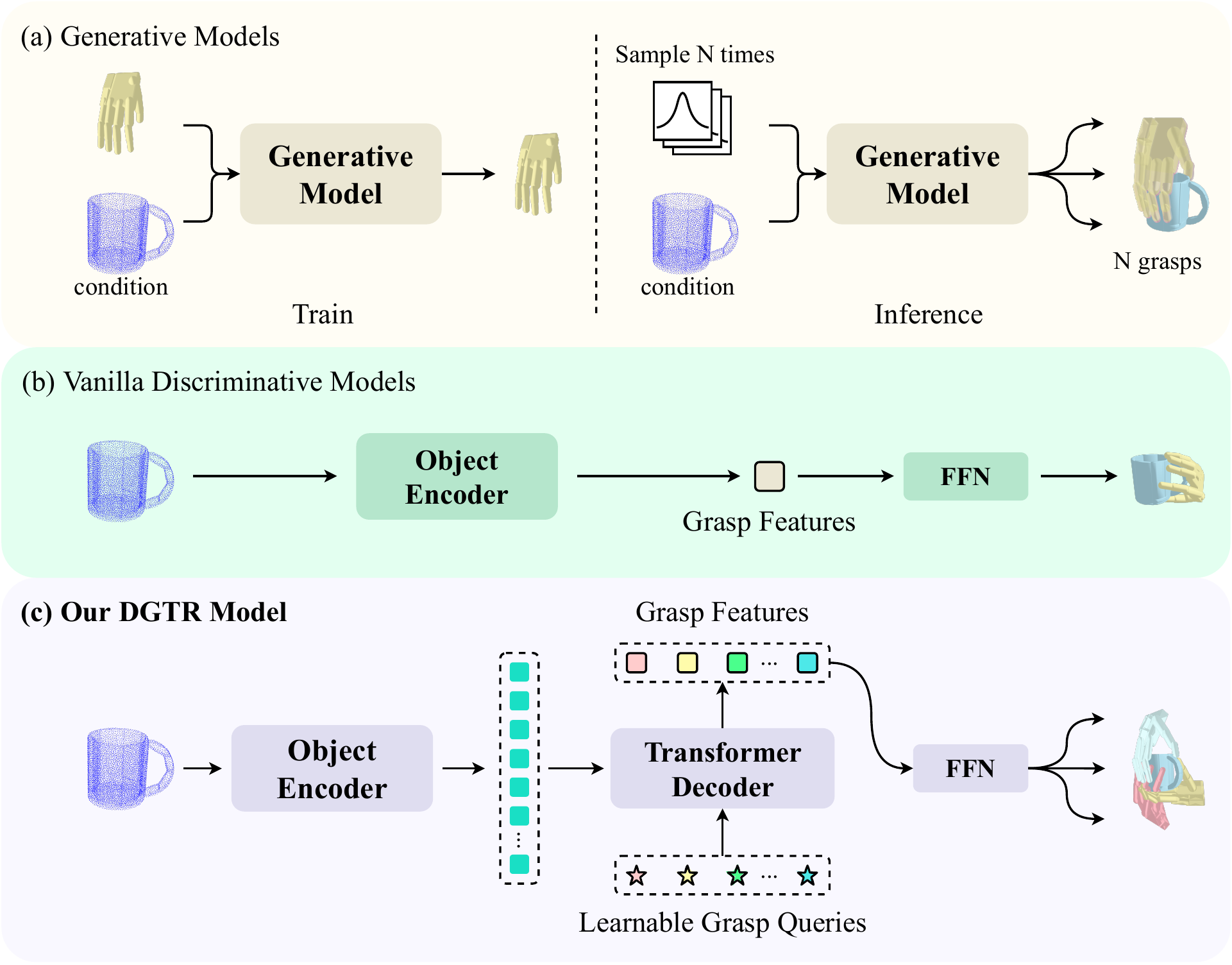}
\caption{
    Comparison of DGTR and other dexterous grasping frameworks. The generative models~(a) usually learn the distribution of the grasp poses conditioned on the object point cloud. At test time, they mainly infer multiple times to generate several grasps but produce nearly identical grasp poses with the same condition. The vanilla discriminative models~(b) mainly learn to predict one grasp pose for the input point cloud.  Our DGTR model~(c) adopts a transformer decoder and learnable queries, and learns to predict a set of diverse grasps poses with one forward pass.
}
\label{figure:comparison}
\end{figure}

With the development of deep learning and large-scale datasets for dexterous grasp generation, learning-based methods achieve considerable performance in grasping quality and generalizability~\cite{grasptta, scene_diffuser, ganhand}. Concurrently, acquiring grasping diversity (especially grasping from various directions) is also a crucial task~\cite{gendexgrasp, unidexgrasp} as it provides the robot with robustness and task flexibility during the manipulation task. Previous learning-based approaches mostly utilize generative models to model the grasp distribution conditioned on the object point cloud as shown in~\Cref{figure:comparison}~(a). However, conditional generative models may consistently generate nearly identical outputs (given the same input) at inference time due to the powerful condition~\cite{condition1, condition2}, except for a diffusion-based model~\cite{scene_diffuser}, which can generate diverse grasps but with low quality. Alternatively, vanilla discriminative models shown in \Cref{figure:comparison}~(b) can only predict a single grasp pose for one input object~\cite{ddg}. Therefore, to obtain diversity, both of them have to rotate the input point cloud and infer multiple times, which is time-consuming and quality-limiting.

In this work, we propose Dexterous Grasp Transformer (DGTR), a novel discriminative framework to tackle the task of predicting diverse and high-quality dexterous grasp poses given the complete object point cloud. We formulate dexterous grasp generation as a set prediction task and design a transformer-based grasping model inspired by the impressive success of Detection Transformers~\cite{detr, 3detr}. As illustrated by \Cref{figure:comparison}~(c), DGTR adopts a transformer decoder and utilizes learnable grasp queries representing different grasping patterns to predict a diverse set of feasible grasp poses by processing the object point cloud only once.

\begin{figure}[t]
    \centering
    \begin{subfigure}[b]{0.32\columnwidth}
      \includegraphics[width=\columnwidth]{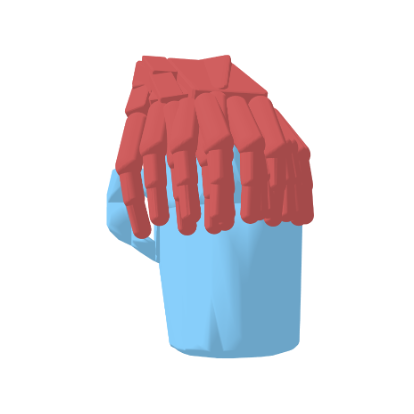}
      \subcaption{$\lambda_{pen} = 500$}
      \label{fig:subfig1}
    \end{subfigure}
    \begin{subfigure}[b]{0.32\columnwidth}
      \includegraphics[width=\columnwidth]{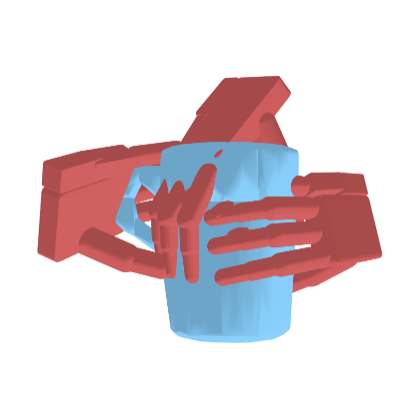}
      \subcaption{$\lambda_{pen} = 0$}
      \label{fig:subfig2}
    \end{subfigure}
    \begin{subfigure}[b]{0.32\columnwidth}
      \includegraphics[width=\columnwidth]{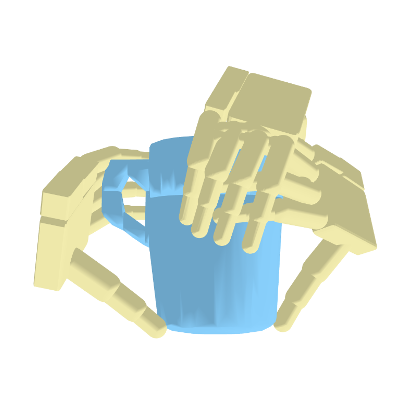}
      \subcaption{Ours}
      \label{fig:subfig3}
    \end{subfigure}
    \caption{
        Comparison of grasp quality and diversity under different penetration loss weights. We visualize 3 grasps for each circumstance. (a) large object penetration weight; (b) zero object penetration weight; (c) our progressive strategies.
    }
    \label{fig:vis_weight}
\end{figure}

However, we observe that DGTR faces an optimization challenge in our task, which results in the dilemma between model collapse and unacceptable object penetration of the predicted grasps. As depicted in~\Cref{fig:vis_weight}~(a), applying a large weight on the object penetration loss causes the model to learn a trivial solution where all predictions are nearly identical. On the contrary, a zero weight for the penetration loss leads to severe object penetration of the grasps, as shown in \Cref{fig:vis_weight}~(b). We identify the main cause of this challenge to be the instability of the Hungarian algorithm, which is exacerbated by the powerful object penetration loss. As the weight of the object penetration loss increases, the matching process becomes more unstable. Consequently, the unstable matching results misguide the optimization process of the model, ultimately causing the model collapse. We conduct abundant analysis and experiments for this in Section \ref{section:DSMT} and \ref{sec:weight}.

To overcome this challenge, we propose progressive strategies for both the training and testing phases, which simultaneously enhance the diversity and quality of grasp poses as demonstrated in \Cref{fig:vis_weight}~(c). Firstly, we present a dynamic-static matching training (DSMT) strategy, which is built on the insight of guiding the model to learn appropriate targets through dynamic matching training and subsequently optimize object penetration through static matching training. This strategy ensures effective optimization of the object penetration loss while directing the model optimization reasonably. Secondly, we present an adversarial-balanced test-time adaptation (AB-TTA) strategy to refine the predicted grasp poses directly in the parameter space of the dexterous hand. Specifically, we utilize a pair of adversarial losses: one repels the hand from the interior of the object, while the other attracts it towards the object's surface. The strategic interaction of the adversarial losses substantially enhances the quality of the grasp and mitigates the penetration. Notably, our AB-TTA neither relies on any 3D mesh information of the objects nor involves complex force analysis or auxiliary models.

Extensive experiments on DexGraspNet dataset show that our methods are capable of generating high-quality and high-diversity grasp poses on thousands of objects. To the best of our knowledge, this is the first work to predict a diverse set of dexterous grasp poses by processing the input object just once, without any need for data preprocessing.

\section{Related Works}
\subsection{Dexterous Grasp Generation}
Dexterous grasping is a promising task as it endows robots with the capability to manipulate objects like humans. Meanwhile, it also presents significant challenges due to the high degree-of-freedom design of dexterous hands. Early methods focus on analytical methods~\cite{analytic_1, DFC, q1} and optimize the hand poses with kinematics and physical mechanisms to a force-closure state. Several works~\cite{dexgraspnet, gendexgrasp} synthesize datasets for dexterous grasps with~\cite{DFC}, but face challenges in the generating speed and success rate.

Recently data-driven methods~\cite{ddgc, unigrasp, efficientgrasp, generating_multi_finger, graspd, toward} have received increasing research attention with the development of deep neural networks. GraspTTA~\cite{grasptta} utilizes a CVAE~\cite{cvae} to synthesize grasps with their hand-object consistency constraints.
UnidexGrasp~\cite{unidexgrasp} proposes two variants of IPDF~\cite{ipdf} and Glow~\cite{glow} to predict object orientation, translation and articulation for the dexterous hand respectively.
Some works~\cite{scene_diffuser, condition_diffusion_1, ganhand} explore conditioned normalizing flow~\cite{glow, normalizing_flow}, generative adversarial network~\cite{gan} and conditioned diffusion models~\cite{rombach2022high} to learn the probabilistic distribution of the dexterous grasps. In contrast, DDG~\cite{ddg} exploits a non-generative model and a differentiable $Q_{1}$ loss to learn one grasp pose for each instance.

However, these methods struggle to generate feasible and diverse grasps given the same input point cloud, either because the condition (\eg, object point cloud) significantly restricts the generation direction of the model, or because of the limitation of the model architecture. To alleviate this problem, our work learns to predict a diverse set of grasps of an object at one time with a transformer-based framework specially designed for dexterous grasp generation.

\begin{figure*}
    \centering
    \includegraphics[width=1.0\textwidth]{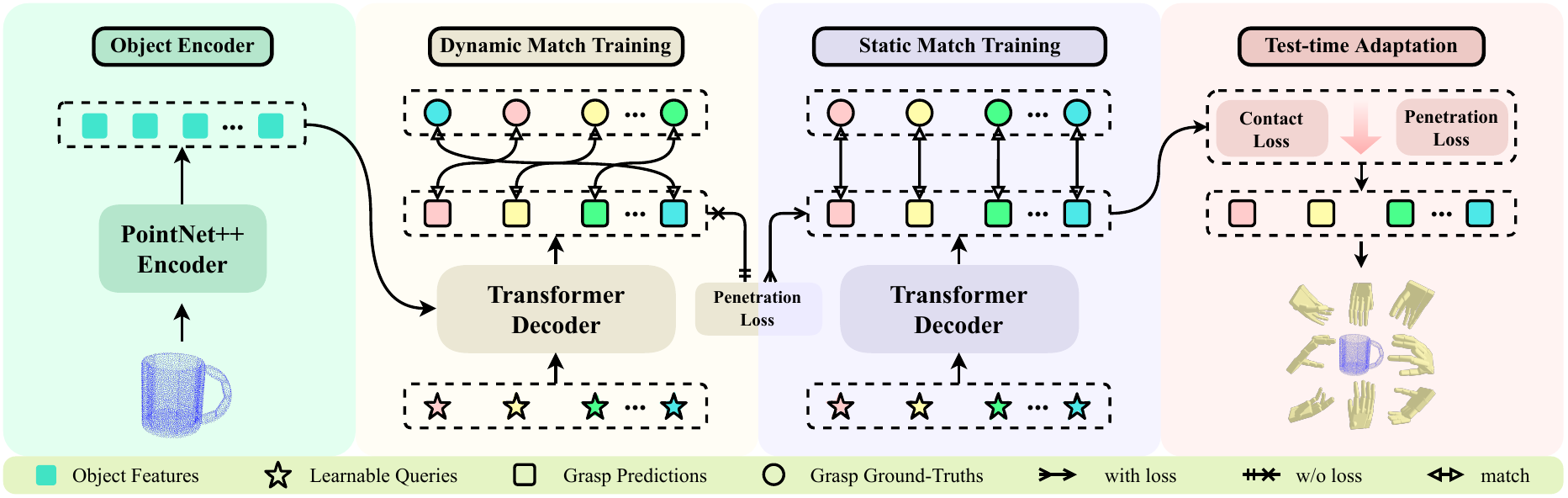}
    \vspace*{-1.5em}
    \caption{\textbf{Overview of our DGTR framework}. The input of DGTR is the complete point cloud $\mathcal{O}$ of an object. First, the PointNet++~\cite{pointnet++} encoder downsamples the point cloud and extracts a set of object features. Next, the transformer decoder takes $N$ learnable query embeddings as well as the object features as input and predicts $N$ diverse grasp poses in parallel.
    In the dynamic matching training stage, our model is trained with the matching result produced by Hungarian Algorithm~\cite{hungarian} and without object penetration loss. In the static matching training stage, we use static matching recorded in the DMT stage to train the model with object penetration loss.
    At test time, we adopt an adversarial-balanced loss to directly finetune the hand pose parameters.}
    \vspace*{-1em}
    \label{fig:overview}
\end{figure*}

\subsection{Vision Transformer}
Vision transformers~\cite{swin, deformable_detr, dynamic_detr, chen2023activating, jiao2023dilateformer, zhou2023twinformer} have received an extensive amount of research attention in recent years, and several of them~\cite{vit, detr, seg_former, point_transformer} introduce novel paradigms for computer vision tasks.
In our work, dexterous grasp generation from a complete point cloud is considered a set prediction task, which is one of the strengths of detection transformers~\cite{detr, 3detr}.
However, conventional detection transformers, which are specially designed for object detection, are unsuited for dexterous grasp generation, because of the absence of supervision for feasible grasps, as well as the optimization challenge arising from the grasp losses.
To tackle this problem, we equip our model with a series of grasp losses for learning diverse and high-quality grasps, and progressive strategies for stable training and penetration optimization.

\section{Dexterous Grasp Transformer}

\subsection{Problem Formulation} \label{formulation}
In this work, we focus on generating high-quality and diverse grasp poses from the complete object point cloud. Specifically, given an object point cloud $\mathcal{O} \in \mathbb{R}^{M \times 3}$ of size $M$, our model learns to generate a set of $N$ dexterous grasp poses $\{\mathbf{g}_{i}\}_{i=1}^{N}$ = $\{(\mathbf{r}_{i}, \mathbf{t}_{i}, \mathbf{q}_{i})\}_{i=1}^{N}$, where $\mathbf{r}_{i} \in \mathbf{SO(3)}$ and $\mathbf{t}_{i} \in \mathbb{R}^{3}$ are the global rotation and translation in the world coordinate, and $\mathbf{q}_{i} \in \mathbb{R}^{J}$ is the joint angles of the $J$-DoF dexterous hand ($J = 22$ for ShadowHand~\cite{shadowhand}).

\subsection{DGTR Architecture}
The model architecture of Dexterous Grasp Transformer (DGTR) contributes most to the diversity and efficiency (\ie $N$ various grasp poses in one forward pass) of our framework. As shown in Figure~\ref{fig:overview}, it mainly consists of three components: 1) a point cloud encoder to extract the object feature, 2) a transformer decoder, and 3) feed-forward networks to predict the grasp poses.
\par
\textbf{Encoder.}
We adopt a three-layer PointNet++~\cite{pointnet++} as the encoder to extract a set of object features. Given an object point cloud $\mathcal{O} \in \mathbb{R}^{M \times 3}$, our encoder outputs the down-sampled point cloud $\mathcal{O}' \in \mathbb{R}^{M'\times 3}$ and the corresponding features $\mathcal{F}' \in \mathbb{R}^{M'\times C'}$.
\par
\textbf{Decoder.}
Inspired by previous set-prediction frameworks~\cite{detr, 3detr}, we cascade Transformer blocks~\cite{transformer} as our decoder to predict an unordered set of grasp poses in parallel. This decoder takes as input the point features $\mathcal{F}'$ and a set of learnable grasping queries $\{q_i\}_{i=1}^N$ to produce grasp features $\{\mathcal{G}_{i}\}_{i=1}^{N}$. Since there is no explicit position information among the point features, we encode the raw points $\mathcal{O}' \in \mathbb{R}^{M' \times 3}$ with an MLP module as the position embedding of encoder features $\mathcal{F}'$.
\par
\textbf{Prediction Heads.}
The grasp pose set $\{\mathbf{g}_{i}\}_{i = 1}^{N}$ = $\{(R_{i}, \mathbf{t}_{i}, \mathbf{q}_{i})\}_{i = 1}^{N}$ are predicted with the final decoder features $\{\mathcal{G}_{i}\}_{i = 1}^{N}$ by three independent MLPs. Both the translation and the joint angle predictions are passed through a sigmoid activation to form a normalized value \wrt the limits of each dimension. And the rotation prediction is normalized to a unit quaternion with the $L2$ normalization.
\par
The unordered predictions are usually matched with their nearest ground truths using the Hungarian Algorithm~\cite{hungarian} before the loss calculation. However, while the Hungarian Algorithm provides an effective solution to train the model regardless of the permutation of the predictions, it also brings ambiguity to the optimizing process of the model, which is a major factor of the dilemma of model collapse and unacceptable object penetration.
We alleviate this problem with a dynamic-static matching training strategy (\Cref{section:DSMT}) and propose an adversarial-balanced loss to further enhance the practicality of the generated grasps at test time (\Cref{section:tta}).

\begin{figure}[t]
    \centering
    \includegraphics[width=0.95\columnwidth]{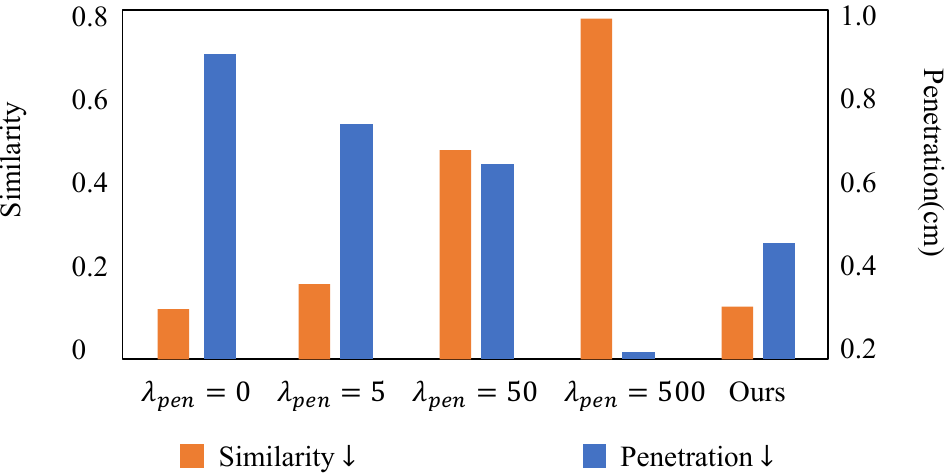}
    \vspace{-0.5em}
    \caption{
        \textbf{Comparative analysis of grasp poses similarity and object penetration with various penetration loss weights.}
        \textit{Similarity} is measured by the cosine similarity of $N$ predicted grasp poses, which represents the \textit{non-diversity}. \textit{Penetration} is the object penetration from the object point cloud to the hand mesh. \textit{Ours} denotes the model trained with our proposed DSMT strategy.
      }
    \vspace{-1em}
    \label{fig:sim}
\end{figure}

\subsection{Dynamic-Static Matching Training Strategy} \label{section:DSMT}
\textbf{Model Collapse \vs Object Penetration.} We discover the optimization challenge when DGTR attempts to learn multiple grasping targets of one object simultaneously. As illustrated in Figure \ref{fig:sim}, DGTR encounters a dilemma between model collapse and the issue of unacceptable object penetration. On one hand, if we impose a heavy penalty on object penetration  (\eg $\lambda_{pen} = 500$), the model tends to be stuck in a trivial solution where it predicts nearly identical grasps for the object. On the other hand, if we reduce this penalty (\eg $\lambda_{pen} = 5$) or even remove it ($\lambda_{pen} = 0$), the predicted grasps suffer from severe object penetration.
\par
\begin{figure}[t]
    \centering
    \includegraphics[width=0.95\columnwidth]{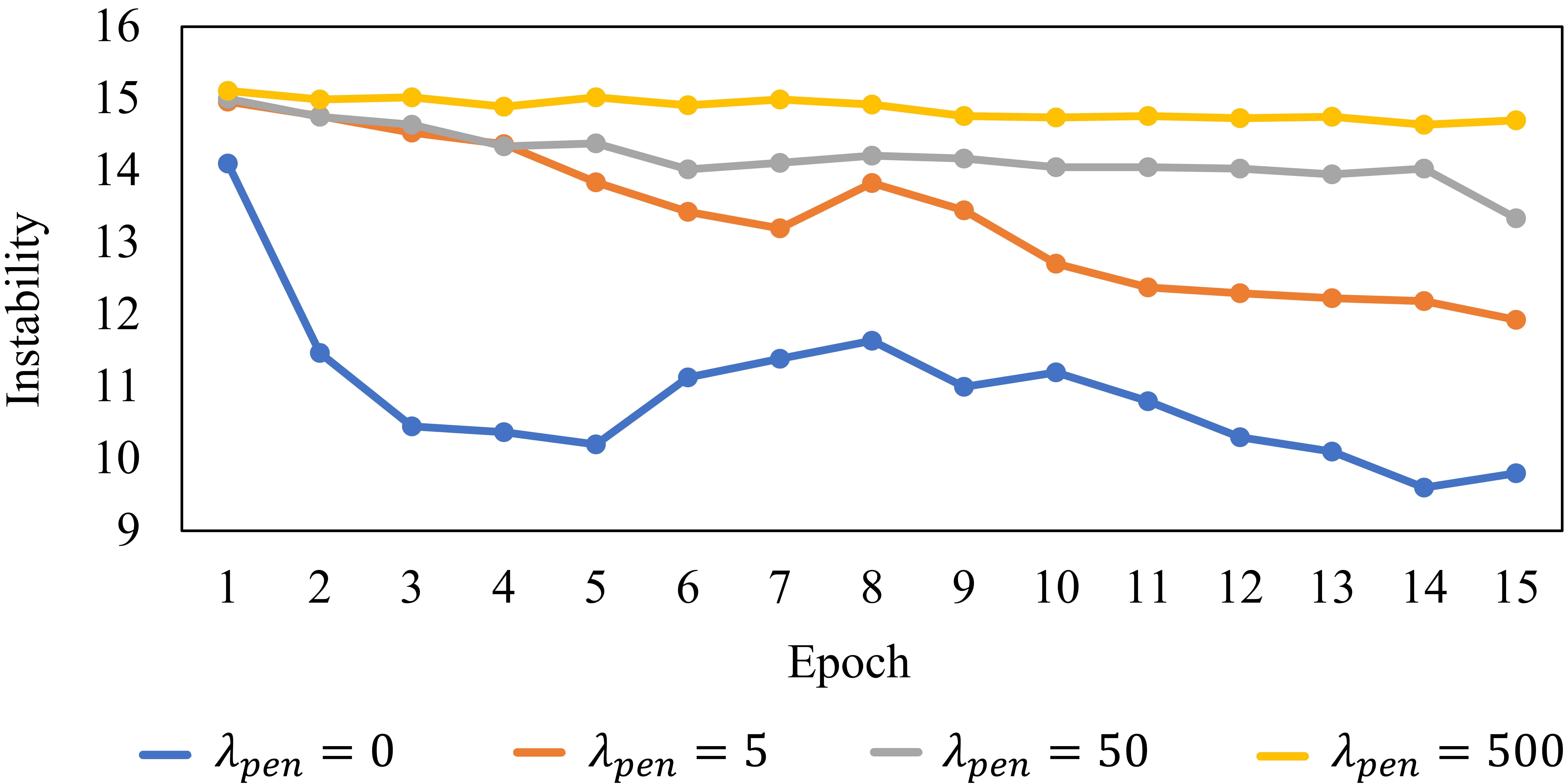}
    \vspace{-0.5em}
    \caption{
        \textbf{Hungarian matching instability during training of different penetration loss weights.} The instability is measured by the \textit{IS} metric introduced in~\cite{dn-detr}, where a higher value indicates greater instability.
    }
    \vspace{-1em}
    \label{fig:hungarian}
\end{figure}

We analyze the reasons why the object penetration penalty could cause model collapse in the case of set prediction. Intuitively, there is a non-trivial gap between the optimizing difficulties of object penetration and hand pose reconstruction. The object penetration loss could be reduced easily by ``pulling'' the hand away from the object. While the latter involves a high-dimensional and non-convex optimization problem, which is inherently difficult to solve. Empirically, the object penetration loss increases the instability of Hungarian Algorithm matching results, which profoundly disturbs the optimizing process. As depicted in Figure \ref{fig:hungarian}, the instability of Hungarian matching increases as $\lambda_{pen}$ becomes larger, which results in ambiguous optimization goals for each query~\cite{fenoaltea2021stable, dn-detr} and eventually causes the model to learn similar grasp poses for all queries.

\textbf{DSMT.}
We serialize the optimizing process and propose a Dynamic-Static Matching Training (DSMT) strategy, aiming to alleviate the optimization challenge arising from the instability of the Hungarian Algorithm and the strong impact of object penetration loss. The key insight is to guide the model learning towards appropriate targets through dynamic training, and subsequently optimizing object penetration through static training.
\par
As illustrated in Algorithm \ref{al:DSMT}, DGTR optimization begins with regular training with the hand regression loss and no object penetration loss for $T_{0}$ epochs (DMT). The matching results between the predictions and the targets are dynamically generated by the Hungarian Algorithm. The learnable queries are adequately trained to learn diverse grasping patterns in this stage.
\par
In the Static Matching Warm-up (SMW) stage, we remove the Hungarian Matching process and utilize fixed and stable matching results recorded in the DMT stage. The objective of this stage is to finetune the model and make it adapt to the given static matching. Thus, we still exclude the object penetration loss in this stage.
\par
In the Static Matching Penetration Training (SMPT) stage, the object penetration loss and the hand-object distance loss~(\cref{eq:van_dist}) are incorporated into the training process. The matching results used in the previous stage are preserved to maintain a stable optimization environment. In this way, the severe penetration issue arising from the lack of object penetration penalty in the previous training stages is significantly alleviated.

\begin{algorithm}[t]
    \renewcommand{\algorithmicrequire}{\textbf{Input:}}
    \renewcommand{\algorithmicensure}{\textbf{Output:}}
    \caption{Dynamic-Static Matching Training}
    \begin{algorithmic}
        \Require
        Object point clouds $\mathcal{O}$, target grasp poses $\mathbf{\hat{g}}$, training epochs $T_{0}$, $T_{1}$, $T_{2}$, and model parameters $\Theta$
        \For{$t = 1$ to $T_{0}$} \Comment{DMT}
        \State $\mathbf{g}_{t} = \Theta(\mathcal{O})$
        \State $\hat{\rho}_{t} = HungarianAlgorithm(\mathbf{g}_{t}, \mathbf{\hat{g}})$
        \State $\mathcal{L}(\mathbf{g}_{t}, \mathbf{\hat{g}}, \hat{\rho}_{t}) = \mathcal{L}_{regress}(\mathbf{g}_{t}, \mathbf{\hat{g}}, \hat{\rho}_{t})$
        \State Update $\Theta$ with $\nabla_{\Theta}\mathcal{L}(\mathbf{g}_{t}, \mathbf{\hat{g}}, \hat{\rho}_{t})$
        \EndFor
        \State $\mathbf{g}_{T_{0}} = \Theta(\mathcal{O})$
        \State $\hat{\rho}_{T_{0}} = HungarianAlgorithm(\mathbf{g}_{T_{0}}, \mathbf{\hat{g}})$
        \For{$t = 1$ to $T_{1}$} \Comment{SMW}
        \State $\mathbf{g}_{t} = \Theta(\mathcal{O})$
        \State $\mathcal{L}(\mathbf{g}_{t}, \mathbf{\hat{g}}, \hat{\rho}_{T_{0}}) = \mathcal{L}_{regress}(\mathbf{g}_{t}, \mathbf{\hat{g}}, \hat{\rho}_{T_{0}})$
        \State Update $\Theta$ with $\nabla_{\Theta}\mathcal{L}(\mathbf{g}_{t}, \mathbf{\hat{g}}, \hat{\rho}_{T_{0}})$
        \EndFor
        \For{$t = 1$ to $T_{2}$} \Comment{SMPT}
        \State $\mathbf{g}_{t} = \Theta(\mathcal{O})$
        \State $\mathcal{L}(\mathbf{g}_{t}, \mathbf{\hat{g}}, \hat{\rho}_{T_{0}}) = \mathcal{L}_{regress}(\mathbf{g}_{t}, \mathbf{\hat{g}}, \hat{\rho}_{T_{0}})$
        \State \hspace{5.3em} $ +~\mathcal{L}_{pen}(\mathbf{g}_{t}, \mathcal{O}) + \mathcal{L}_{van-dist}(\mathbf{g}_{t}, \mathcal{O})$
        \State Update $\Theta$ with $\nabla_{\Theta}\mathcal{L}(\mathbf{g}_{t}, \mathbf{\hat{g}}, \hat{\rho}_{T_{0}})$
        \EndFor
        \Ensure Optimized model parameters $\Theta$
    \end{algorithmic}
    \label{al:DSMT}
\end{algorithm}

\subsection{Adversarial-Balanced Test-Time Adaptation} \label{section:tta}
\textbf{Object Contact \vs Object Penetration.} To further improve the practicality of the predicted grasps, we propose an adversarial-balanced test-time adaptation (AB-TTA) strategy to refine the predicted grasps during the test phase. It is worth noting that our AB-TTA eliminates the need for complex force analysis or auxiliary models. Specifically, this strategy mainly minimizes a pair of adversarial losses, the object penetration loss $\mathcal{L}_{pen}$ and hand-object distance loss $\mathcal{L}_{dist}$ in the parameter space of the dexterous hand.
However, the comprehensive optimization of these two losses is challenging. The penetration loss can be easily reduced (\ie, pulling the hand away from the object) in the parameter space without appropriate constraints, causing the hand-object distance loss to lose efficacy.
Hence, we incorporate two key designs to facilitate a balanced decrease of these adversarial losses, which brings considerable improvement in both hand-object contact and hand-object penetration.
\par
\textbf{AB-TTA.}
Our AB-TTA is based on the perception that the generated grasp poses are already or nearly valid, only requiring slight adjustments.
Firstly, we propose to moderate the displacement of the global translation of the root link of the dexterous hand during the optimization process by downscaling its gradient with $\beta_{t}$. Moderating the global translation constrains the over-optimization of object penetration loss, which promotes the effectiveness and stability of the adaptation.
\par
Secondly, we present a generalized tta-distance loss to address the ineffectiveness of vanilla distance loss used in ~\cite{dexgraspnet}. The vanilla distance loss is defined as:
\begin{equation}
\setlength{\abovedisplayskip}{5pt}
\setlength{\belowdisplayskip}{5pt}
    \mathcal{L}_{van-dist} = \sum_{i}\mathbb{I}(d(p_{i}) < \tau) * d(p_{i}),
\label{eq:van_dist}
\end{equation}
where $\mathbb{I}(\cdot)$ is the indicator function, $\tau$ is a  contact threshold to filter out the outliers, and $d(p_{i})$ is the distance between the nearest point on the object point cloud and the $i^{th}$ keypoint $p_i$ on the predicted hand. We observe that the vanilla distance loss will be 0 if the hand is too far away from the object, where no point meets the conditions $(d(p_{i}) < \tau)$. As a result, the hand is unlikely to be ``pushed'' towards the object again since the distance loss has been 0.
We improve the hand-object distance loss by defining a more general condition which constrains the hand keypoints that initially touched the object to remain in contact during optimization. The generalized tta-distance loss is defined as:
\begin{equation}
\setlength{\abovedisplayskip}{5pt}
\setlength{\belowdisplayskip}{5pt}
    \mathcal{L}_{tta-dist} = \sum_{i}{\mathbb{I}((d(p^{c}_{i}) < \tau) \lor (d(p^{r}_{i}) < \tau)) * d(p^{r}_{i})},
\end{equation}
where $p^{c}_{i}$ and $p^{r}_{i}$ are the $i^{th}$ keypoints of the initial coarse hand and the refined hand at the current iteration, respectively. As a result, a input hand which is nearly valid would not be pulled too far away from the object.
\par
In addition, due to the high DoF of dexterous hands, we also add self-penetration loss $\mathcal{L}_{spen}$ in AB-TTA. Thus, the overall loss function for AB-TTA is
\begin{equation}
\setlength{\abovedisplayskip}{5pt}
\setlength{\belowdisplayskip}{5pt}
    \mathcal{L}_{ab-tta}
    = \alpha_{1} * \mathcal{L}_{pen} + \alpha_{2} * \mathcal{L}_{tta-dist} + \alpha_{3} * \mathcal{L}_{spen}.
\end{equation}
The details of all losses are in \Cref{grasploss} and \textit{Appendix} A.

\begin{figure*}[t]
    \centering
    \includegraphics[width=1.0\textwidth]{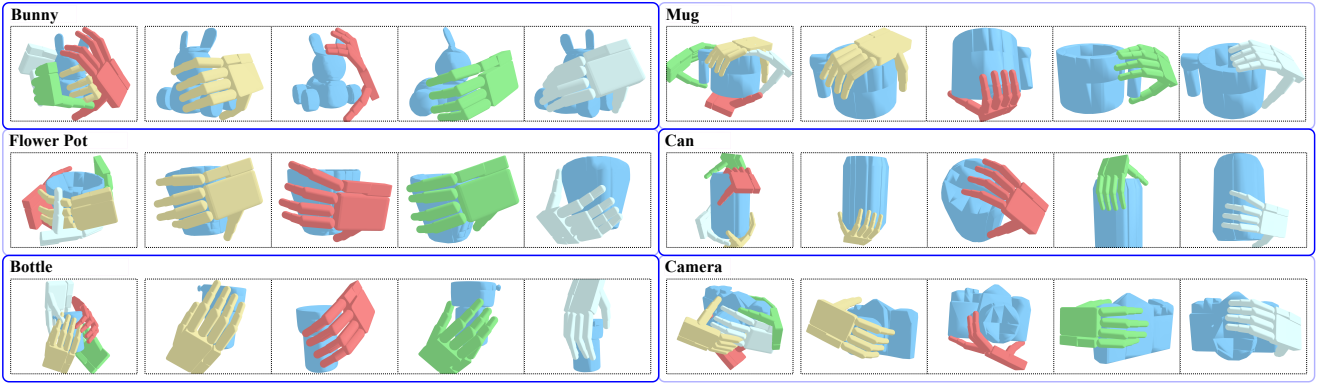}
    \caption{Visualization of predicted dexterous hand poses. We visualize four grasp poses in five images for each object. The first image visualizes all grasps together to demonstrate their global positions. The following four images mainly visualize the details of the grasp pose. These visualization results qualitatively indicate that the proposed DGTR framework is capable of generating diverse and feasible grasps with the same input and only \textbf{in one forward pass}. More visualization results can be found in \textit{Appendix} C.}
    \label{fig:vis_outputs}
    \vspace{-1em}
\end{figure*}

\subsection{Grasp Losses}
\label{grasploss}
The optimization of DGTR involves the grasp losses and the bipartite matching between the predictions and the ground truths. We denote the $i^{th}$ predicted item as $\mathbf{x}_{i}$ and the $j^{th}$ ground-truth item as $\mathbf{\hat{x}}_{j}$ ($\mathbf{x} \in \{\mathbf{g}, \mathbf{t}, \mathbf{r}, \mathbf{q} \}$) in the following paragraphs of this section.
\par
\textbf{Hand Parameters Regression Loss}. We utilize the smooth $L1$ loss~\cite{fast_rcnn} as $\mathcal{L}_{trans}$ and $\mathcal{L}_{joints}$ to regress the translations and joint angles. For the rotation, we maximize the similarity of the predicted and ground-truth quaternions with
$\mathcal{L}_{rotation}(\mathbf{r}_{i}, \mathbf{\hat{r}}_{j}) = 1.0 - \lvert \mathbf{r}_{i} \cdot \mathbf{\hat{r}}_{j} \rvert$,
where $(\cdot)$ is the inner product operation.
The overall regression loss for hand parameters is a weighted sum of the above losses:
\vspace{-1mm}
\begin{equation}
\begin{aligned}
&\mathcal{L}_{param}(\mathbf{g}_{i}, \mathbf{\hat{g}}_{j})
= \lambda_1 * \mathcal{L}_{trans}(\mathbf{t}_{i}, \mathbf{\hat{t}}_{j}) \\
&+ \lambda_2 * \mathcal{L}_{joints}(\mathbf{q}_{i}, \mathbf{\hat{q}}_{j})
+ \lambda_3 * \mathcal{L}_{rotation}(\mathbf{r}_{i}, \mathbf{\hat{r}}_{j}).
\end{aligned}
\vspace{-1mm}
\end{equation}
\par
\textbf{Hand Chamfer Loss}. We incorporate a hand chamfer loss $\mathcal{L}_{chamfer}(\mathbf{g}_{i}, \mathbf{\hat{g}}_{j})$ to explicitly minimize the discrepancies between the actual shapes of the predicted and ground-truth hands. Specifically, we apply $\mathbf{g}_{i}$ and $\mathbf{\hat{g}}_{j}$ to the dexterous hand and obtain the hand meshes $\mathcal{H}(\mathbf{g}_{i})$ and $\mathbf{\mathcal{H}(\hat{g}}_{j})$ by forward kinematics. Then we sample the hand point clouds $\Phi(\mathbf{g}_{i})$ and $\Phi(\mathbf{\hat{g}}_{j})$ from the corresponding meshes and calculate the Chamfer distance~\cite{chamfer} between them.
\par
\textbf{Penetration Loss}. We employ two penetration losses: 1) $\mathcal{L}_{pen}(\mathbf{g}_{i}, \mathcal{O})$~\cite{unidexgrasp}: object penetration calculated by the signed squared distance function from object point cloud to the hand mesh, and 2) $\mathcal{L}_{spen}(\mathbf{g}_{i})$~\cite{dexgraspnet}: self penetration depth from the keypoints of the hand to themselves.
\par
\textbf{Cost Function for Bipartite Matching.}
To obtain a bipartite matching between predictions and ground truths, the cost function for each pair of $(\mathbf{g}_i, \mathbf{\hat{g}}_j)$ is defined as:
\begin{align}
    \mathcal{C}(\mathbf{g}_i, \mathbf{\hat{g}}_j)
    &= \omega_1 * \mathcal{L}_{trans}(\mathbf{t}_i, \mathbf{\hat{t}}_j)
    + \omega_2 * \mathcal{L}_{joints}(\mathbf{r}_i, \mathbf{\hat{r}}_j) \notag \\
    &+ \omega_3 * \mathcal{L}_{rotation}(\mathbf{q}_i, \mathbf{\hat{q}}_j).
\end{align}
Let $\rho \in \mathcal{P}_{N}$ be a permutation of $N$ elements, and assume that $K = M = N$. We utilize Hungarian Matching Algorithm~\cite{hungarian} to compute the optimal assignment $\hat{\rho}$:
\begin{equation}
\setlength{\abovedisplayskip}{3pt}
\setlength{\belowdisplayskip}{3pt}
    \hat{\rho} = \mathop{argmin}\limits_{\rho \in \mathcal{P}_{N}}\sum_{i}^{K} \mathcal{C}(\mathbf{g}_{i}, \mathbf{\hat{g}}_{\rho_{i}}).
\end{equation}
The process of computing $\hat{\rho}$ when $M \neq N$ is similar, except that we leave the redundant predictions or ground truths unmatched. As a result, there are $K = min\{M, N\}$ matched pairs accounting for the overall loss.
\par
\textbf{Overall Loss Function.}
The overall grasp loss for DGTR training is a weighted sum of the aforementioned losses, which is formulated as:
\begin{align}
    \mathcal{L}_{grasp}(\mathbf{g}_{i}, \mathbf{\hat{g}}_{\rho_{i}}, \mathcal{O})
    &= \mathcal{L}_{param}(\mathbf{g}_{i}, \mathbf{\hat{g}}_{\hat{\rho}_{i}}) \notag \\
    &+ \lambda_{4} * \mathcal{L}_{chamfer}(\mathbf{g}_{i}, \mathbf{\hat{g}}_{\hat{\rho}_{i}}) \\
    &+ \lambda_{5} * \mathcal{L}_{spen}(\mathbf{g}_{i})
    + \lambda_{6} * \mathcal{L}_{pen}(\mathbf{g}_{i}, \mathcal{O}). \notag
\end{align}
This loss is averaged among all matched pairs $\{\mathbf{g}_{i}, \mathbf{\hat{g}}_{\hat{\rho}_{i}}\}_{i = 1}^{K}$. 

\section{Experiments}

\begin{table*}[t]
    \footnotesize
    \setlength{\tabcolsep}{12.38pt}
    \centering
        \begin{tabular}{c|*{5}{c}|*{3}{c}}
        \toprule
        \multirow{2}*{Method} & \multicolumn{5}{c|}{Quality} & \multicolumn{3}{c}{Diversity} \\
        & $Q_{1}\uparrow$ & $\eta_{np}\uparrow$ & $\eta_{tb}\uparrow$ & $\eta_{success} \uparrow$ & $Pen. \downarrow$ & $\delta_{t}\uparrow$ & $\delta_{r}\uparrow$ & $\delta_{q}\uparrow$ \\
        \midrule
        GraspTTA~\cite{grasptta} & 0.0271 & 18.95 & 15.90 & 24.5 & 0.678 & 8.09 & 7.53 & 7.90 \\
        UniDexGrasp~\cite{unidexgrasp} & 0.0462 & \textbf{97.29} & 50.94 & 37.1 & 0.121 & 9.64 & 7.49 & 29.29 \\
        SceneDiffuser~\cite{scene_diffuser} & 0.0129 & 96.21 & 22.88 & 25.5 & \textbf{0.107} & \textbf{54.84} & \textbf{52.27}  & \textbf{39.75} \\
        DGTR (ours) & \textbf{0.0515} & 75.78 & \textbf{69.62} & \textbf{41.0} & 0.421 & 47.77 & 51.66 & 27.81 \\
        \midrule
        DDG~\cite{ddg} & 0.0582 & 84.53 & 56.63 & \textbf{67.5} & \textbf{0.173} & 6.25 & 6.25 & 6.25 \\
        DGTR\textsuperscript{*} (ours) & \textbf{0.0921} & \textbf{99.51} & \textbf{81.28} & 66.6 & 0.313 & \textbf{19.66} & \textbf{20.68} & \textbf{15.11} \\
        \bottomrule
        \end{tabular}
    \vspace{-0.5em}
    \caption{Results on DexGraspNet\cite{dexgraspnet} compared with the state-of-the-art \textbf{in one forward pass} condition. DGTR\textsuperscript{*} is a practical variant of DGTR concentrating on grasp quality. Note that DDG~\cite{ddg} is not in the same setting as ours and serves as a quality reference here.}
    \label{table:sota}
\end{table*}

\begin{table}[t]
    \footnotesize
    \centering
    \setlength{\tabcolsep}{4.5pt}
    \vspace{-1em}
    \begin{tabular}{c|cc|*{4}{c}}
    \toprule
    Method & $n_{pass}$ & $n_{grasp}$ & $T_{inf}$ (ms) $\downarrow$ & $\delta_{t}\uparrow$ & $\delta_{r}\uparrow$ & $\delta_{q}\uparrow$ \\
    \midrule
    Uni.~\cite{unidexgrasp} & 1 & 16 & 58.3 $\pm$ 4.1 & 9.64 & 7.49 & 29.29 \\
    Uni.~\cite{unidexgrasp} & 4 & 4 & 153.7 $\pm$ 8.8 & 18.37 & 22.20 & 36.36 \\
    Uni.~\cite{unidexgrasp} & 16 & 1 & 530.6 $\pm$ 12.2 & 25.04 & 44.31 & \textbf{38.65} \\
    Ours & 1 & 16 & \textbf{20.4 $\pm$ 3.3}  & \textbf{47.77} & \textbf{51.66} & 27.81 \\
    \bottomrule
    \end{tabular}
    \vspace{-0.5em}
    \caption{Comparison with multiple pass methods. $T_{inf}$ is the total time to generate all grasp poses. $n_{pass}$ refers to the times of object's point cloud being rotated and passed to the grasping model. $n_{grasp}$ is the number of grasp poses generated per pass. }
    \label{table:muti-view}
\end{table}

\begin{table}
    \centering
    \footnotesize
    \setlength{\tabcolsep}{9.4pt}
    \begin{tabular}{c|cccc}
        \toprule
        Method & $Q_{1}\uparrow$ & $Pen. \downarrow$ & $\eta_{np}\uparrow$ & $\eta_{tb}\uparrow$ \\
        \midrule
        DGTR       & 0.0515 & 0.421 & 75.78 & 69.62 \\
        w/o AB-TTA & 0.0278 & 0.466 & 52.36 & 65.10 \\
        w/o DSMT   & 0.0115 & 0.869 & 7.69 & 96.84  \\
        \bottomrule
    \end{tabular}
    \vspace{-0.5em}
    \caption{Effectiveness of each component of DGTR.}
    \vspace{-1.5em}
    \label{table:ab}
    \end{table}

\subsection{Dataset and Evaluation Metrics}
We evaluate the proposed DGTR framework in the challenging dexterous grasping benchmark DexGraspNet~\cite{dexgraspnet}, which contains 1.32 million grasps of ShadowHand~\cite{shadowhand} for 5355 objects from more than 133 object categories. The official training-validation split is used in our experiments.
\par
We use five metrics to conduct comprehensive evaluations of the generating quality of DGTR. That is,
1) \textbf{Mean $Q_{1}$}~\cite{q1} reflects grasp stability. We follow~\cite{dexgraspnet} to set the contact threshold to 1cm and set the penetration threshold to 5mm. 2) \textbf{Maximal penetration depth (cm)} ($Pen.$), which is the maximal penetration depth from the object point cloud to hand meshes. 3) \textbf{Non-penetration ratio $\eta_{np}$ (\%)}, which is the proportion of the predicted hands with a maximal penetration depth of less than 5mm. 4) \textbf{Torque balance ratio $\eta_{tb}$ (\%)}, denoting the percentage of torque-balanced grasps (\ie $Q_{1} > 0$). 5) \textbf{Grasping success rate $\eta_{success}$ (\%) in Isaac Gym}~\cite{isaac}. Following ~\cite{dexgraspnet}, we consider a grasp pose valid if the grasp can hold the object steadily under any one of the six gravity directions.
\par
For diversity, we introduce the new metrics, 6) \textbf{occupancy proportion} of translations $\delta_{t}$, rotations $\delta_{r}$ and joint angles $\delta_{q}$~(\%),  to quantitatively measure the ability of a model to grasp objects from a diverse range of directions, orientations, and joint angles. Generally, we discretize the continuous parameter space into $\xi = 16$ uniform bins and calculate the proportion of occupied spaces for different grasps of each object. For $\delta_{t}$, we uniformly sample $\xi$ points as the bins on a unit sphere with Fibonacci sampling, and then assign each grasp to a bin based on the cosine similarity between its global translation and the corresponding direction of the point. For $\delta_{r}$ and $\delta_{q}$, we discretize the range of Euler angle into $\xi$ bins. Intuitively, higher values of $\delta_{t}$ indicate that the predicted grasps can move to more areas of the object and grasp it from more directions, while higher $\delta_{r}$ and $\delta_{q}$ suggest more various hand orientations and gestures. All details of metrics can be found in \textit{Appendix A}.

\subsection{Implementation Details}
Our DGTR is implemented with PyTorch~\cite{paszke2019pytorch} and trained on a single RTX 4090 GPU. The number of queries $N$ is set to 16.
The training epochs for each stage in DSMT are $T_{0} = 15$, $T_{1} = 5$, $T_{2} = 5$. We set $\omega_{1} = 2.0$, $\omega_{2} = 1.0$, and $\omega_{3} = 2.0$ for the Hungarian Algorithm cost function. During DMT and SMW, the loss weight are $\lambda_{1} = 10.0$, $\lambda_{2} = 10.0$, $\lambda_{3} = 10.0$, $\lambda_{4} = 1.0$, $\lambda_{5} = 10.0$, $\lambda_{6} = 0.0$. In the SMPT stage, $\lambda_{6}$ is set to $50.0$, and distance loss weight is $10.0$. For AB-TTA, we set $\beta_{t}=0$, $\alpha_{1} = 5$, $\alpha_{2} = 3$, and $\alpha_{3} = 5$. More details can be found in \textit{Appendix} A.

\begin{table*}[t]
    \parbox{.37\textwidth}{
        \centering
        \footnotesize
        \setlength{\tabcolsep}{2.3pt}
        \begin{tabular}{c|*{4}{c}}
            \toprule
             Method & $Q_{1}\uparrow $ & $Pen. \downarrow$ & $\eta_{np}\uparrow$ & $\eta_{tb}\uparrow$ \\
            \midrule
            DMT & 0.0115 & 0.869 & 7.69 & 96.74 \\
            DMT + SMW & 0.0100 & 0.879 & 6.55 & 97.25 \\
            \rowcolor{gray!25} 
            DMT + SMW + SMPT&0.0278 & 0.466 & 52.36 & 65.10 \\
            \hline
            w/o Static & 0.0064 & 0.600 & 36.84 & 56.67 \\
            w/o Warm & 0.0271 & 0.482 & 50.03 & 67.15 \\
            \bottomrule
        \end{tabular}
        \vspace{-0.5em}
        \caption{
            Ablation study for three stages in DSMT. \textit{Static} and \textit{Warm} are static matching and warm-up for SMPT. The complete DSMT is colored in \textcolor{gray}{gray}.
        }
        \label{table:DSMT}
    }
    \hfill
    \parbox{.27\textwidth}{
        \centering
        \footnotesize
        \setlength{\tabcolsep}{4.5pt}
        \scalebox{1.0}{
        \begin{tabular}{c|*{4}{c}}
            \toprule
            $N$ & $Q_{1}\uparrow $ & $\delta_{t}\uparrow$ &$\delta_{r}\uparrow$ & $\delta_{q}\uparrow$ \\
            \midrule
            4 & 0.0392 & 18.40 & 21.85 & 9.60 \\
            8 & 0.0305 & 28.26 & 33.64 & 12.96 \\
            16 & 0.0278 & 47.77 & 51.66 & 27.81 \\
            32 & 0.0275 & 72.13 & 65.88 & 19.48 \\
            64 & 0.0170 & 89.57 & 78.41 & 25.50 \\
            \bottomrule
        \end{tabular}}
        \vspace{-0.5em}
        \caption{Analysis of the Number of grasp queries. $N$ is the number of queries.}
        \label{table:query}
    }
    \hfill
    \parbox{.30\textwidth}{
        \centering
        \footnotesize
        \setlength{\tabcolsep}{3.8pt}
        \begin{tabular}{c|cccc}
            \toprule
            $\lambda_{pen}$ & $Q_{1}\uparrow$ & $Pen. \downarrow$ & $\eta_{np}\uparrow$ & $\eta_{tb}\uparrow$ \\
            \midrule
            0  & 0.0115 & 0.869 & 7.69 & 96.84 \\
            5  & 0.0203 & 0.717 & 22.94 &84.79 \\
            50 & 0.0109 & 0.662 & 36.76 & 60.62\\
            500 & 0.0020 & 0.207 & 78.19 &16.75 \\
            $0\rightarrow50$ & 0.0061 & 0.651 & 31.45 & 59.86\\
            \bottomrule
        \end{tabular}
        \vspace{-0.5em}
        \caption{Analysis of different object penetration loss weight. $\lambda_{pen}$ is the penetration weight.}
        \label{table:weight}
    }
\end{table*}

\subsection{Dexterous Grasp Generation Performance}
\subsubsection{Comparison with SOTA in one forward pass}
We first compare SOTA dexterous grasp generation methods with DGTR in our setting, where each method is allowed to infer once. DDG~\cite{ddg} takes multi-view images as input and only predicts one grasp pose for each object, which serves as a quality reference. SceneDiffuser~\cite{scene_diffuser}, GraspTTA~\cite{grasptta} and UniDexGrasp~\cite{unidexgrasp} samples 16 times in a batch, with the same object point cloud as condition.
\par
The evaluation results are shown in \Cref{table:sota}. For grasp quality, DGTR surpasses the SOTA generative models in several important metrics. Note that UniDexGrasp has remarkable performance in $\eta_{np}$ and $Pen.$ but with a low $\eta_{tb}$, which suggests low contact with the object, while DGTR has a more balanced performance and higher success rate.
Moreover, owing to the capability of generating diverse grasps, DGTR can efficiently select top-4 results (DGTR*) by the number of contact points and object penetration during inference without extra inputs. In this scenario, DGTR\textsuperscript{*} has comparable results with DDG~\cite{ddg}.
\par
For diversity, DGTR surpasses UniDexGrasp~\cite{unidexgrasp} and GraspTTA~\cite{grasptta} by a large gap in terms of $\delta_{t}$ and $\delta_{r}$, which indicates that DGTR is able to grasp the object from a variety of directions. SceneDiffuser~\cite{scene_diffuser} has higher diversity but with much lower quality. More comparisons with SceneDiffuser are in \textit{Appendix B}.
The results demonstrate that DGTR achieves overall SOTA performance and excels in generating high-quality and diverse grasps.
\par
\begin{figure}[t]
    \centering
    \vspace{-1em}
    \begin{subfigure}[b]{0.3\columnwidth}
        \centering
        \includegraphics[width=0.5\columnwidth]{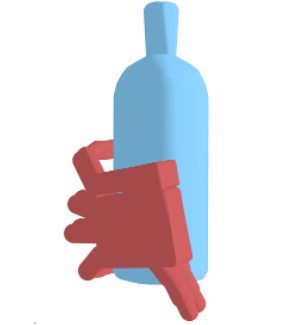}
        \subcaption{GraspTTA~\cite{grasptta}}
        \label{subfig:grasptta}
    \end{subfigure}
    \hfill
    \hfill
    \hfill
    \begin{subfigure}[b]{0.3\columnwidth}
        \centering
        \includegraphics[width=0.5\columnwidth]{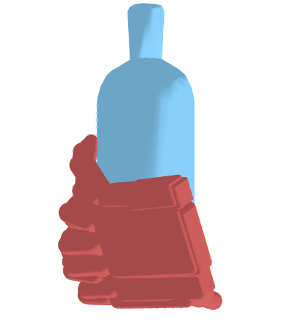}
        \subcaption{UniDexGrasp~\cite{unidexgrasp}}
        \label{subfig:unidexgrasp}
    \end{subfigure}
    \hfill
    \hfill
    \hfill
    \begin{subfigure}[b]{0.3\columnwidth}
        \centering
        \includegraphics[width=0.5\columnwidth]{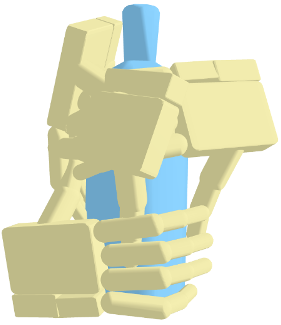}
        \subcaption{Ours}
        \label{subfig:ours}
    \end{subfigure}
    \vspace{-0.5em}
    \caption{
        Comparison of grasp diversity in one forward pass with 4 outputs. The diversity of our DGTR significantly surpasses \cite{grasptta} and \cite{unidexgrasp} in one forward pass.
    }
    \label{fig:vis_compare}
    \vspace{-1em}
\end{figure}

We visualize the predicted grasp poses of several objects in \Cref{fig:vis_outputs} to provide a qualitative result of DGTR. DGTR is capable of generating high-quality grasps of an object from various directions with different poses in one forward pass.
Furthermore, \Cref{fig:vis_compare} highlights the diversity of DGTR in comparison to two other generative methods.

\vspace{-1em}
\subsubsection{Comparison with SOTA in multiple forward pass}
\Cref{table:muti-view} presents a comparison of grasping diversity and inference time between UniDexGrasp in multiple forward passes and DGTR in one forward pass. UniDexGrasp first utilizes a probabilistic model to sample rotations and then rotates object point clouds to generate grasps in multiple passes. As shown in~\Cref{table:muti-view}, DGTR exhibits significantly lower time consumption compared to the multi-pass UniDexGrasp. More importantly, DGTR outperforms UniDexGrasp with 16 forward passes in $\delta_{t}$ and $\delta_{r}$. This indicates that DGTR offers more diverse grasping hand positions and enables grasping from a wider range of directions.

\subsection{Ablation Study}
\subsubsection{Dynamic-Static Matching Training Strategy}
\label{exp:msts}
As demonstrated in \Cref{table:ab}, our DSMT significantly enhances $Q_{1}$ by 3.5 times, while reducing $Pen.$ by nearly $50\%$. \Cref{table:DSMT} provides more details on the performance after each training stage (DMT, SMW and SMPT) in DSMT. The results highlight the critical role of static matching, which optimizes the model  towards the proper direction and significantly reduces object penetration.

\subsubsection{Adversarial-Balanced Test-Time Adaptation}
\begin{figure}[t]
    \vspace{-1.5em}
    \centering
    \includegraphics[width=0.85\columnwidth]{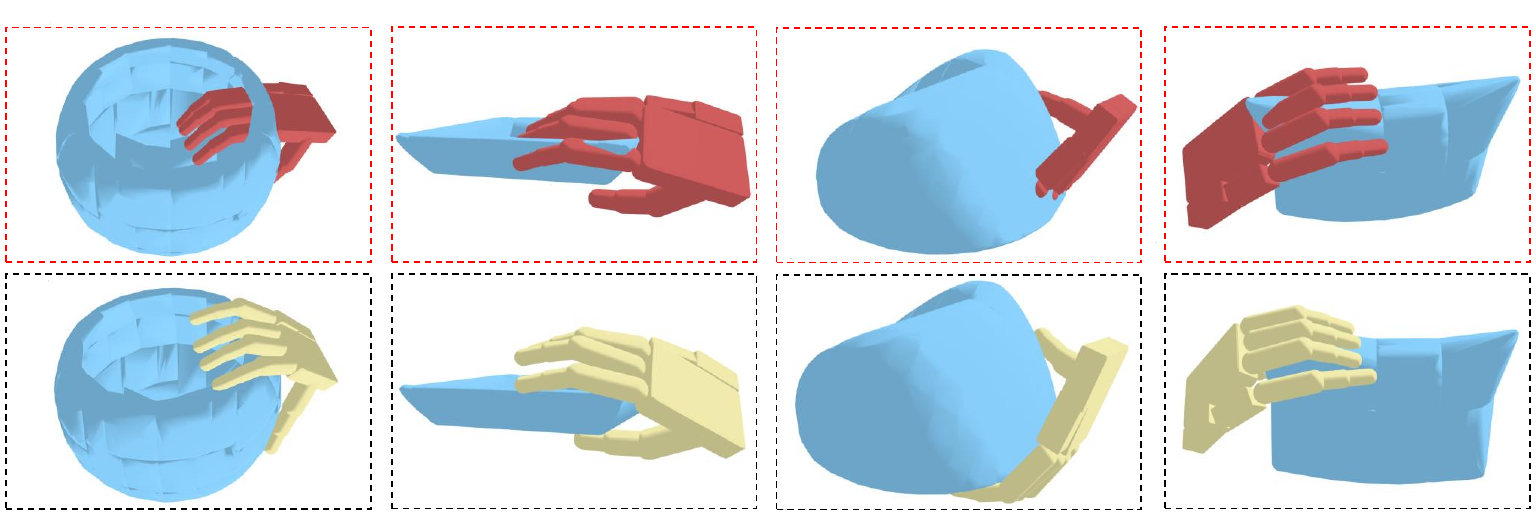}
    \caption{
        Visualization of grasps \textcolor[RGB]{139, 0, 0}{\textbf{before}} and \textcolor[RGB]{204, 204, 0}{\textbf{after}} AB-TTA.
    }
    \vspace{-1em}
    \label{fig:vis_tta}
\end{figure}

We conduct ablation studies on our AB-TTA module and the results are in \Cref{table:ab}, \Cref{table:tta}, and \Cref{fig:vis_tta}. As shown in \Cref{table:ab}, our AB-TTA significantly increases $Q_{1}$ by 1.85-fold, and enhances $\eta_{np}$, and $\eta_{tb}$ at the same time.
Furthermore, \Cref{table:tta} shows that the integration of our key designs (\ie, generalized tta-distance loss (GDis) and translation moderation strategy (TM)) are indispensable, while the simple implementation of TTA (\ie, penetration and vanilla distance loss (VDis)) has limited effect.
Furthermore, our AB-TTA module demonstrates superior grasp quality compared to ContactNet-TTA~\cite{grasptta}, and it can even boost the $Q_{1}$ and $\eta_{tb}$ performance of ContactNet-TTA.

\begin{table}
\footnotesize
\centering
\vspace{-1.3em}
\setlength{\tabcolsep}{6.5pt}
\begin{tabular}{*{5}{c}|*{3}{c}}
    \toprule
    Pen& VDis& GDis &TM & CN &
    $Q_{1}\uparrow$  & $\eta_{np}\uparrow$ & $\eta_{tb}\uparrow$ \\
    \midrule
    $\checkmark$&$\checkmark$& & & & 0 & 100 & 0 \\
    $\checkmark$& &$\checkmark$ & & & 0.0125 & 77.15 & 28.08 \\
    $\checkmark$ &  & &$\checkmark$ & & 0.0295 & 75.31 & 48.56 \\
    $\checkmark$& & & &$\checkmark$ & 0.0435 & 98.54 & 50.50 \\
    $\checkmark$& & $\checkmark$ & $\checkmark$ &$\checkmark$
    & 0.0491 & 78.24 & 64.80 \\
    \rowcolor{gray!25}
    $\checkmark$& & $\checkmark$ & $\checkmark$ &  & 0.0515 & 75.78 & 69.62 \\
    \bottomrule
\end{tabular}
\caption{
    Ablation study for designs in AB-TTA. \textit{Pen} and \textit{Dis} denote penetration and distance loss. \textit{GDis} and \textit{TM} are our generalized tta-distance loss and translation moderation strategy. \textit{CN} refers to ContactNet~\cite{grasptta}. Our whole AB-TTA is colored in \textcolor{gray}{gray}.
}
\vspace{-1em}
\label{table:tta}
\end{table}

\subsection{DGTR Analysis}
We conduct a series of analytic experiments for DGTR. We discuss object penetration weight and the number of queries below. Please refer to the \textit{Appendix} B for more analysis.

\subsubsection{Loss Weight for Object Penetration}
\label{sec:weight}
The results in \Cref{table:weight} show that the object penetration decreases as $\lambda_{pen}$ increase, but a severe non-contact issue occurs concurrently. As illustrated in \Cref{fig:sim} and \Cref{fig:hungarian}, the instability of Hungarian matching leads to model collapse when we apply a large penetration loss. And it is worth noting that gradually increasing $\lambda_{pen}$ from 0 to 50 after several warm-up epochs cannot tackle this problem ($\lambda_{pen}=0\rightarrow50$ in \Cref{table:weight}). We believe that learning to predict multiple grasps simultaneously is a more difficult optimization process compared to the previous one-to-one grasping learning. And our proposed progressive strategies (\ie, DSMT and AB-TTA) tackle this challenge effectively.

\subsubsection{Number of Grasping Queries}
We conduct experiments to analyze the effect of the number of grasping queries. As shown in ~\Cref{table:query}, the grasp quality $Q_{1}$ tends to decrease as the number of queries increases, which suggests that simultaneous learning a larger set of grasping poses is a challenge.
Furthermore, the diversity increases as the number of queries becomes larger, implying that DGTR can learn a more diverse set of grasp with a greater number of queries.

\section{Conclusions}
In this work, we propose DGTR (Dexterous Grasp Transformer), a novel discriminative framework for dexterous grasp generation. Our progressive strategies, including dynamic-static matching training (DSMT) strategy and adversarial-balanced test-time adaptation (AB-TTA), substantially improve grasping stability and reduce penetration. To the best of our knowledge, DGTR is the first work to introduce set prediction formulation into dexterous grasp domain and achieves both high quality and diversity with one forward pass. We believe that DGTR holds good development potential in robotic dexterous grasping scenarios, such as task-oriented and real-world dexterous grasp generation.

\section*{Acknowledgements}
We thank Jialiang Zhang for his helpful discussion. This work was supported in part by the National Key Research and Development Program of China (2023YFA1008503), NSFC (U21A20471, U1911401), and Guangdong NSF Project (No. 2023B1515040025, 2020B1515120085).

{
    \small
    \bibliographystyle{ieeenat_fullname}
    \bibliography{main}
}

\end{document}

%% file: preamble.tex
\usepackage[dvipsnames]{xcolor}
